\documentclass[runningheads]{llncs}

\usepackage{eccv}

\usepackage{eccvabbrv}

\usepackage{graphicx}
\usepackage{booktabs}

\usepackage[accsupp]{axessibility}  

\usepackage{multirow}
\usepackage{makecell}
\usepackage{scalerel} 
\usepackage{enumitem} 
\usepackage{pifont} 
\newcommand{\xmark}{\ding{55}}%
\usepackage[dvipsnames]{xcolor} 
\usepackage{fontawesome}
\usepackage{marvosym}
\usepackage{pifont}
\usepackage{subcaption} 
\usepackage{soul}

\usepackage{color, colortbl}
\definecolor{Pink}{rgb}{1,0.941,1.}
\definecolor{PositiveGreen}{rgb}{0,0.56,0}

\usepackage{hyperref}

\usepackage{orcidlink}

\begin{document}

\title{
Keypoint Promptable Re-Identification
}


\author{Vladimir Somers\inst{1, 2, 3}\orcidlink{0000-0001-5787-4276} \and
Alexandre Alahi\inst{1}\orcidlink{0000-0002-5004-1498} \and
Christophe De Vleeschouwer\inst{2}\orcidlink{0000-0001-5049-2929}}

\authorrunning{V. Somers et al.}


\institute{Ecole Polytechnique Fédérale de Lausanne (EPFL), Switzerland \and 
Université Catholique de Louvain (UCLouvain), Belgium \and 
Sportradar, Switzerland}


\maketitle


\newcommand\model{KPR}
\newcommand\OccludedPT{Occluded PoseTrack-ReID}
\newcommand\OccPT{Occ-PTrack}
\newcommand\randocc{BIPO}
\newcommand\randocclusions{Batch-wise Random-Occlusions}
\newcommand\G{G}
\newcommand\ffg{f_{fg}}
\newcommand\fgl{f_{gl}}
\newcommand\fcc{f_{cc}}
\newcommand\Y{Y}
\newcommand\scores{M} 
\newcommand\A{A}
\newcommand\F{F}
\newcommand\glb{hol}
\newcommand\lcl{part}
\newcommand\reid{ReID}



                    
\begin{abstract}
Occluded Person Re-Identification (ReID) is a metric learning task that involves matching occluded individuals based on their appearance.
While many studies have tackled occlusions caused by objects, multi-person occlusions remain less explored.
In this work, we identify and address a critical challenge overlooked by previous occluded {\reid} methods: the Multi-Person Ambiguity (MPA) arising when multiple individuals are visible in the same bounding box, making it impossible to determine the intended ReID target among the candidates.
Inspired by recent work on prompting in vision, we introduce Keypoint Promptable ReID (KPR), a novel formulation of the ReID problem that explicitly complements the input bounding box with a set of semantic keypoints indicating the intended target.
Since promptable re-identification is an unexplored paradigm, existing {\reid} datasets lack the pixel-level annotations necessary for prompting.
To bridge this gap and foster further research on this topic, we introduce {\OccludedPT}, a novel {\reid} dataset with keypoints labels, that features strong inter-person occlusions.
Furthermore, we release custom keypoint labels for four popular ReID benchmarks.
Experiments on person retrieval, but also on pose tracking, demonstrate that our method systematically surpasses previous state-of-the-art approaches on various occluded scenarios.
Our code, dataset and annotations are available at \url{https://github.com/VlSomers/keypoint_promptable_reidentification}.
\end{abstract}

\section{Introduction} \label{section:intro}


Person re-identification ({\reid}) \cite{Ye2021} is a challenging retrieval task that involves matching a query image of a person of interest with other person images. 
{\reid} finds wide-ranging applications in multi-object tracking \cite{GHOST}, pedestrian flow analysis \cite{Alahi2017}, sport understanding \cite{deepsportradarv1, soccernet22, Istasse2022, PRTreID, sngamestate} and video-surveillance \cite{market1501}. 
However, {\reid} is a difficult task due to various factors including inaccurate bounding boxes, luminosity changes, poor image quality, and occlusions.


\begin{figure}[t!]
    \centering
    \begin{subfigure}[b]{0.49\textwidth} 
        \centering
        \includegraphics[width=\textwidth]{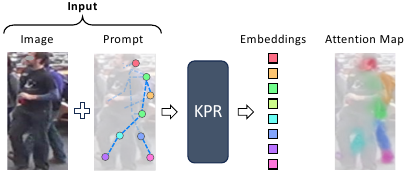}
        \caption{Keypoint Promptable ReID (KPR)}
        \label{fig:pull_figure_1}
    \end{subfigure}
    \\
    \begin{subfigure}[b]{0.49\textwidth} 
        \centering
        \includegraphics[width=.8\textwidth]{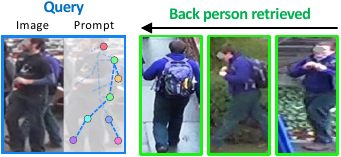}
        \caption{Back (occluded) person prompted}
        \label{fig:pull_figure_2}
    \end{subfigure}
    \hfill 
    \begin{subfigure}[b]{0.49\textwidth} 
        \centering
        \includegraphics[width=.8\textwidth]{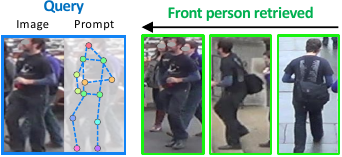}
        \caption{Front person prompted}
        \label{fig:pull_figure_3}
    \end{subfigure}
    
    \caption{Overview of our proposed Keypoint Promptable ReID (KPR) method.
    KPR takes an image with keypoints prompts as input and produces part-based features of the prompted target.
    The prompt instructs the model to focus on a specific individual, i.e. the back blue jacket man (b) or the front black t-shirt man (c) in this example.
    Colored dots illustrate the positive keypoints prompt, with one color per body part.
    }
    \label{fig:pull_figure}
\end{figure}

\begin{figure}[t!]
\centering
  \begin{subfigure}{.49\textwidth} 
    \centering
    \includegraphics[height=2.3cm,keepaspectratio]{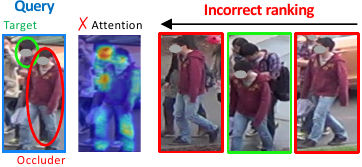}
    \captionsetup{justification=centering} 
    \caption{Existing ReID models (e.g. BPBreID \cite{bpbreid}) are confused by \textbf{MPA}}
    \label{fig:related_work_1}
  \end{subfigure}
  \hfill 
  \begin{subfigure}{.49\textwidth} 
    \centering
    \includegraphics[height=2.3cm,keepaspectratio]{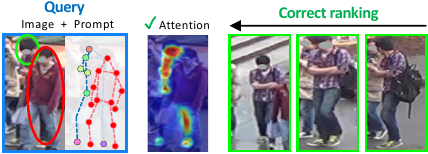}
    \captionsetup{justification=centering} 
    \caption{
    KPR explicitly addresses MPA with positive (pastel) and negative (red) keypoints prompt
    }
    \label{fig:related_work_2}
  \end{subfigure}

  \caption{
        Person retrieval with Multi-Person Ambiguity (\textbf{MPA}).
        Green/red borders are correct/incorrect matches.
        Red/pastel dots indicate negative/positive prompts.
  }
  \label{fig:related_work}
\end{figure}

In recent years, there has been a growing interest in addressing the occluded re-identification task \cite{occluded_reid_survey}, where the {\reid} target might be occluded by various objects or other people.
In this work, we identify and explicitly address a critical challenge that has been overlooked by previous occluded {\reid} methods \cite{FED, bpbreid}: the presence of \textbf{Multi-Person Ambiguity} (MPA) inherent to bounding box annotations.
The multi-person ambiguity arises when a bounding box image contains multiple individuals, making it challenging even for humans to accurately identify the intended {\reid} target among all the candidates.
\Cref{fig:related_work_1} shows how MPA can negatively impact the re-identification model, causing feature mixing among different individuals or even a focus on an unintended person. 

%
To tackle this multi-person ambiguity, it is necessary to incorporate additional pixel-level information, such as keypoints or segmentation masks, provided by an upstream human operator or vision model.
This additional data plays a vital role in assisting the downstream {\reid} method to distinguish the intended target from other candidates.

Inspired by recent advances in promptable\footnote{A ``prompt" typically refers to an input instruction that guides the model's response. In vision tasks, e.g. segmentation, prompts can be any kind of data that specifies what the model should focus on \cite{SAM}.} methods for vision transformers (e.g., SAM \cite{SAM}), we propose to explicitly use additional semantic\footnote{Semantic keypoints are distinct points within a person image that carry semantic information about specific body parts, typically obtained with pose estimation \cite{coco}. } keypoints as inputs to disambiguate images with multiple individuals.
Semantic keypoints are suitable for two crucial use cases: 1) manual prompting by a human operator with a few clicks on an image, and 2) automated prompting by an upstream pose estimation model. 
Nevertheless, our architecture could be easily extended to support segmentation masks as prompts.

Prior research \cite{PFD, PVPM, HOReID} on \textit{pose-guided methods} has explored the use of additional information from external pose estimation models.
However, none of these works explicitly tackled the multi-pedestrian ambiguity issue.

Following above insights, we propose \textbf{{\model}}, a novel \textit{\textbf{K}eypoint-\textbf{P}romptable transformer for part-based person \textbf{R}e-identification}, illustrated in \cref{fig:pull_figure_1}.
{\model} is a transformer-based model that takes an image along with \textit{semantic keypoints} as input.
It then produces \textit{part-based features}, each representing a distinct body part of the {\reid} target, along with their respective visibility scores.
Part-based methods have demonstrated superior performance \cite{ISP, bpbreid} in re-identifying occluded individuals because they utilize only the visible parts for comparison.
Our method can process both \textit{positive} and \textit{negative} keypoints, which respectively represent the target and non-target pedestrians.
As illustrated in \cref{fig:pull_figure_2}, prompts can be used to re-identify the desired (front or back) person in case of occlusion.
Furthermore, {\model} is designed to be \textit{prompt-optional} to offer more practical flexibility.
This means the same model can be used without prompt on non-ambiguous images, or with prompt when dealing with occlusions, while consistently achieving state-of-the-art performance in both cases.
To demonstrate the advantages of keypoint prompts for re-identification, we evaluate {\model} on two tasks: \textit{person retrieval} and \textit{multi-person pose tracking}.

Moreover, since no current {\reid} dataset is provided with explicit target identification in case of multi-person ambiguity, we introduce a new dataset called \textbf{\textit{{\OccludedPT}}}, nicknamed ``\textit{{\OccPT}}", and derived from the PoseTrack21 dataset \cite{Doering2022}.
This dataset offers high-quality keypoints annotations and features a substantial number of images with multi-person occlusions.


Overall, we summarize the main contributions of our work as follows:
\setlist{nolistsep}
\begin{enumerate}[noitemsep]

\item We propose \textit{{\model}}, a novel prompt-optional model for part-based ReID. 
To our knowledge, we are the first to introduce a visual prompting mechanism for person re-identification, and the first to explicitly address the ambiguity caused by multi-person occlusions.
\item We introduce \textit{{\OccPT}}: the first multi-person occluded {\reid} dataset with explicit target identification through manual keypoints annotations. 
Furthermore, we propose new keypoint annotations for four popular re-identification datasets. 

\item Our method outperforms all previous state-of-the-art approaches on the occluded {\reid} task, with a notable \textit{+12.6\% mAP} and \textit{+9.2\% Rank-1} improvement on the well-established Occluded-Duke \cite{PGFA} benchmark.

\end{enumerate}

Our ambition is to advocate for a shift from the ambiguous bounding-box approach to {\reid} and promote a keypoint-centric paradigm.
To this end, we publicly release our codebase, proposed dataset, and annotations, to provide a common setup for evaluating promptable {\reid} methods.

\section{Related Work} \label{section:related_work}
\textbf{Multi-Person Ambiguity}:
A previous study \cite{FED} addressed a related issue called ``Non-Target Pedestrians (NTP)," similar to our concept of ``Multi-Person Ambiguity" (MPA).
However, their focus was on mitigating the negative impact of NTP during training.
In contrast, we provide a method to explicitly disambiguate the intended {\reid} target during both training and testing phases.

\textbf{Pose-guided {\reid}}:
We detail here the fundamental difference between our novel keypoint promptable approach and the traditional pose-guided methods.
Although none of the previous works explicitly address the multi-person ambiguity (MPA) issue, we identify two categories within these pose-guided approaches: 1) those inherently unable to address MPA \cite{bpbreid, VGTri, PGFA, PVPM, PIRT, MoS, OPReID, OAMN}, and 2) those with the potential to overcome MPA \cite{HOReID, PFD}.
In the first category, methods like BPBreID \cite{bpbreid} or Pirt \cite{PIRT} utilize pose labels only during training, and can not leverage additional information at test time to disambiguate multiple persons.
Moreover, methods relying on horizontal stripes (VGTri \cite{VGTri}, PGFA \cite{PGFA}) or the affinity fields of a pose model (PVPM \cite{PVPM}), are meant to tackle object-occlusions, but struggle to disambiguate multiple persons. 
In the second category, methods like HOReID \cite{HOReID} and PFD \cite{PFD} directly employ keypoints from a single person, enabling focus on a specified target.
However, HOReID \cite{HOReID} is restricted to utilizing spatial features at the exact keypoints locations, thus overlooking critical appearance cues on entire body regions.
Finally, PFD \cite{PFD}, and all previous pose-guided methods including HOReID, incorporate pose information \textit{after appearance encoding has occurred}, e.g., for local pooling of a spatial feature map, but do not leverage pose data to guide the appearance encoding process as our method do.
Indeed, our approach implements a \textit{prompt-aware appearance encoding}, since it conditions feature extraction on the input keypoints.
This enables \textit{better feature disentanglement between the target and the occluders} from the early feature extraction stages, resulting in more accurate representations.
Furthermore, our method is \textit{prompt-optional}, setting it apart for its flexibility from these previous pose-guided methods.
Finally, our method can ingest additional \textit{negative keypoints}, to further disambiguate a multi-person occlusion scenario.

\textbf{Prompting in Vision}:
Recent works have introduced promptable transformers to tackle object segmentation. 
For instance, SAM \cite{SAM} employs two networks: a transformer encoder to generate generic spatial features and a lightweight promptable decoder to integrate user inputs for interactive segmentation. 
In contrast, our approach directly prompts a feature encoder, so that \textit{feature extraction is explicitly driven by the prompts}. 
Unlike SimpleClick \cite{SimpleClick}, which prompts a transformer with a single 2D point and a previous mask for iterative segmentation, our model ingests multiple keypoints concurrently, enhancing body part localization and representation by \textit{leveraging their semantic information}.



\section{Methodology} \label{section:methodology}
\begin{figure*}[t!]
\centering
\includegraphics[width=0.99\linewidth]{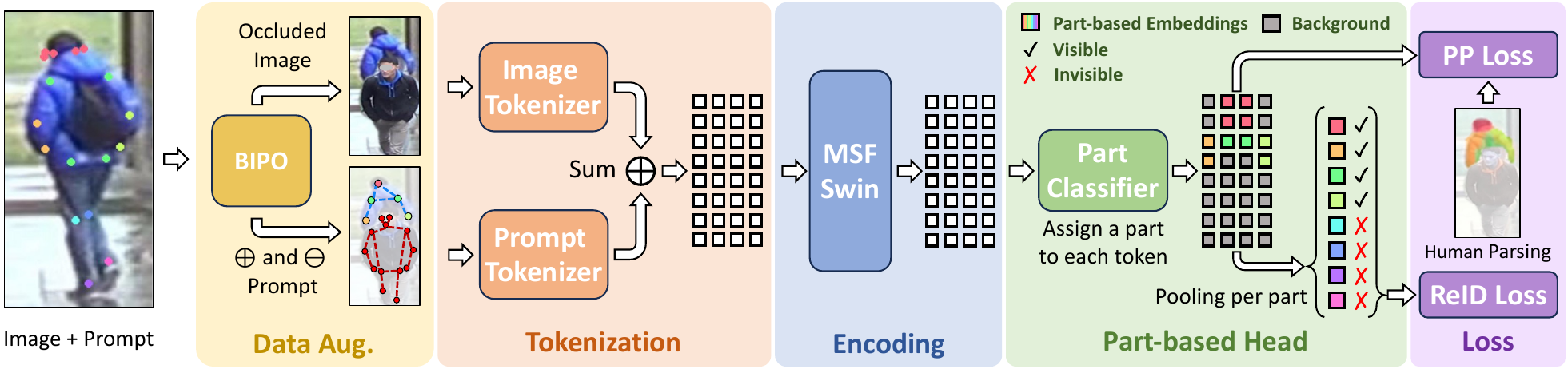}
  \caption{Architecture overview of our proposed Keypoint Promptable ReID ({\model}) model. 
  The Batch-wise Inter-Person Occlusion (BIPO) data augmentation is first applied to generate artificial inter-person occlusions (\cref{section:bipo}).
  The image and the optional positive/negative ($\oplus / \ominus$) prompts are then tokenized and summed (\cref{section:tokenization}).
  Tokens are then fed to our proposed Multi-Stage feature Fusion (MSF) Swin transformer to generate high-resolution feature maps (\ref{section:encoding}).
  The feature map is then fed to a Part-based Head (PBH), which assigns a part (or the background) to each token with a part classifier, and then averages all tokens of the same part to produce the final $K$ part-based embeddings \{$f_1$, ..., $f_K$\} and their binary visibility scores \{$v_1$, ..., $v_K$\}, with $v_i \in \{0, 1\}$, and visually denoted by \{\xmark, \checkmark\} (\cref{section:pbh}).
  A part $i$ with no token assigned is considered invisible (i.e., $v_i = 0$), and ignored when computing two samples' similarity.
  {\model} is illustrated here for K=8, with a unique color for each body part: \{\textit{head}, \textit{torso}, \textit{right/left arm}, \textit{right/left leg}, and \textit{right/left feet}\}.
  Finally, the entire pipeline is trained with two losses: a Part-Prediction (PP) Loss and a ReID Loss (\cref{section:training}). 
  }
\label{fig:architecture}
\end{figure*}

The overall architecture of our model {\model} is illustrated in \cref{fig:architecture} and is based on a Swin transformer backbone \cite{Swin}.
The next subsections first describe the various components of the architecture.
We then present the training process, including our proposed data augmentation.
Finally, we describe the inference procedure and conclude with an intuitive discussion of our design choices.

\subsection{Tokenization} \label{section:tokenization}
The tokenization procedure coverts raw image and keypoints information into $C_i$-dimensional tokens, to be subsequently processed by the Swin encoder.

\textbf{Image tokenization}:
The input image, with dimensions $H \times W$, undergoes tokenization using the conventional Swin patch embed module. 
This process generates $H_t \times W_t = \frac{H}{4} \times \frac{W}{4}$ spatial tokens of dimensions $C_i$ by embedding each $4 \times 4$ image patch through a linear layer.

\textbf{Keypoints tokenization}:
As previously mentioned, our model uses two types of semantic keypoints as prompts: \textit{positive keypoints} indicating the {\reid} target and \textit{negative keypoints} indicating other individuals.
We organize these prompt keypoints into K+1 groups: the first group contains all negative keypoints, and the remaining K groups contain subsets of the positive keypoints.
Each subset of positive keypoints actually corresponds to a semantic body region. 
For instance, with $K=8$, we define groups like: \{\textit{head}, \textit{torso}, \textit{right/left arm}, \textit{right/left leg}, and \textit{right/left feet}\}.
For each group $i \in$ \{1, ..., $K+1$\}, a unique heatmap \(M_i\) of size $H \times W$ is created by drawing a Gaussian 2D kernel at each of its keypoint's center location ($x$, $y$).
The kernel's standard deviation is set to a fraction $\alpha_{G}$ of the image width. 
Invisible keypoints are ignored.
These $K+1$ heatmaps are then concatenated into a global spatial tensor \(M\) of size \(H \times W \times (K+1)\).
Finally, each \(4 \times 4\) patch in \(M\) is linearly projected to produce prompt tokens of dimension \(C_i\). 
Like the image tokenizer, the keypoints tokenizer outputs therefore $H_t \times W_t$ spatial tokens of dimension $C_i$.

\textbf{Token fusion}:
Images and keypoints tokens are fused by summing tokens at the same spatial location, producing $H_t \times W_t$ spatial tokens with dimension $C_i$. 
These tokens are then forwarded to the Swin-based encoder. 

\textbf{Design motivations}:
Prompted keypoints are organized into $K+1$ channels to preserve semantic meaning before tokenization.
It allows the model to differentiate positive and negative prompts, and to distinguish prompt information of different body parts, for better part localization.
\textit{Prompt-aware appearance encoding}: Moreover, feeding the network with keypoint information before the encoding stage is critical in enabling the network to disentangle the {\reid} target features from other persons' features.
Finally, this design offers the flexibility to support segmentation mask prompts by replacing $M$ with a double-channel map of positive and negative segmentation masks.
Regarding token fusion, the sum operation maintains the input dimension for the Swin backbone, permitting the use of pre-trained weights from the image-only model. 
\textit{Prompt optionality}: This design also offers the flexibility of running the model without prompts, by processing the image tokens alone, with no architectural adjustments required.


\subsection{MSF Swin Encoder} \label{section:encoding} 
A \textit{Swin-transformer} \cite{Swin} is employed as the backbone feature extractor for its great performance in vision tasks and its ability to capture high-frequency details, which is beneficial to {\reid} \cite{PHAZhang}.
Moreover, we propose to enhance Swin with a Multi-Stage feature Fusion strategy, denoted as ``MSF", to output feature maps with a high spatial resolution.
Indeed, similar to CNN, each Swin stage reduces the number of tokens by a factor of 4 while doubling the channel dimension.
To obtain a high-resolution token map, the outputs of all stages are bilinearly upsampled to the same resolution $H_t \times W_t$ as the output of the tokenization module, then concatenated into a single feature map along the channel dimension, and finally passed through a linear layer to obtain the final spatial token map of size $H_t \times W_t \times C_o$.
Different from the standard Swin, our MSF Swin therefore preserves the input tokens resolution.
As pointed out in prior works \cite{BoT, ISP}, a high-resolution feature map enriches the granularity of the features and brings significant performance improvements (see \cref{table:main_ablation_study}, Exp. 3).


\subsection{Part-based Head} \label{section:pbh} 

The encoder yields $H_t \times W_t$ tokens, which are sent to the Part-based Head (PBH) module, as illustrated in \cref{fig:architecture}. 
Each token is first labeled either as one of the $K$ body parts or as background, by a \textit{token-wise part classifier}.
The part classifier is implemented as a linear layer with $K+1$ output logits followed by the softmax operation, and is applied individually on each token.
The assignment of each token to one of the $K$ body parts (or the background) is therefore formulated as a classification task with $K+1$ classes. 
We term the probability map of size $H_t \times W_t \times (K+1)$ produced by this part classifier over all tokens as ``\textit{part-attention maps}".
These maps are showcased as colored heatmaps in \cref{fig:pull_figure_1} and \cref{fig:qualitative_results_2}, with one color per channel (the background channel is ignored). 
Finally, tokens having the same assigned body part are averaged together, resulting in $K$ part-based embeddings \{$f_1$, ..., $f_K$\} of size $C_o$.
Background tokens are ignored.
Parts with no tokens assigned are considered invisible and given a visibility score $v_i=0$.
The corresponding invisible embedding $f_i$, with $v_i=0$, is therefore ignored for downstream computation.
The PBH module thus produces $K$ part-based embeddings \{$f_1$, ..., $f_K$\} and their binary visibility scores \{$v_1$, ..., $v_K$\}.

\textbf{Human parsing labels}: To train the token-wise part classifier, each spatial token is assigned a ground-truth class, i.e., part label, derived from the coarse human parsing labels provided by \cite{bpbreid}.
These human parsing labels are illustrated on the right of \cref{fig:architecture} and in \cref{fig:bipo}.


\subsection{Training Process} \label{section:training}

The overall objective function used to train the model is:
{\small{
\begin{equation} \label{overall_loss}
L = L_{\rm{ReID}} + \lambda_{pp} L_{pp}\ ,
\end{equation}
}}
where $L_{ReID}$ is the ReID loss supervised with identity labels, $L_{pp}$ is the token-wise part-prediction loss, and $\lambda_{pp}$ is a weight parameter empirically set to $0.3$.

\textbf{Part Prediction Loss}:
The token-wise part classifier introduced in \cref{section:pbh} is supervised with a Part Prediction (PP) Loss that is actually a cross-entropy loss applied on each spatial token as in \cite{bpbreid}.

\textbf{Re-identification loss}:
As a ReID objective, we employ the GiLt Loss \cite{bpbreid}, that is specifically designed to train part-based {\reid} models. 
It has the benefit of being robust to occlusions and non-discriminative body parts.
GiLt combines an identity loss \cite{BoT} and a batch-hard triplet loss \cite{triplet}.
The distance function employed for the triplet loss is the average of all local part-based distances.


\subsection{BIPO Data Augmentation} \label{section:bipo}

Since most {\reid} datasets feature limited occlusion in their training set, a model often struggles to handle the occlusions it encounters during testing.
To address this, we propose a novel Batch-wise Inter-Person Occlusion data augmentation (BIPO)\footnote{Inspired by the copy-paste technique for segmentation \cite{copy-paste}.}, to generate artificial inter-person occlusions on training images, prompts, and human parsing labels.
When provided with a training sample, BIPO randomly selects a different person's image (the occluder) from the same training batch, contours it with a segmentation mask derived from the human parsing labels, and overlays it on the main image.
The human parsing label and keypoints prompt of the training image are then updated accordingly.
Finally, all positive keypoints from the occluder are added to the training prompt to serve as additional negative points.

\begin{figure}[t!]
\centering
    \begin{subfigure}{.38\textwidth} 
    \centering
    \includegraphics[height=3.6cm,keepaspectratio]{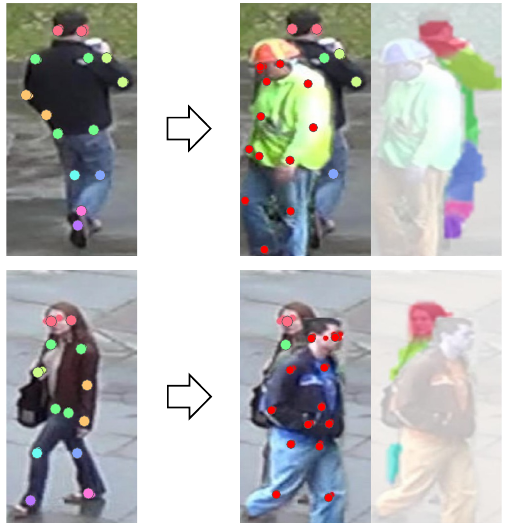}
    \caption{BIPO}
    \label{fig:bipo}
  \end{subfigure}
    \begin{subfigure}{.59\textwidth} 
    \centering
    \includegraphics[height=3.6cm,keepaspectratio]{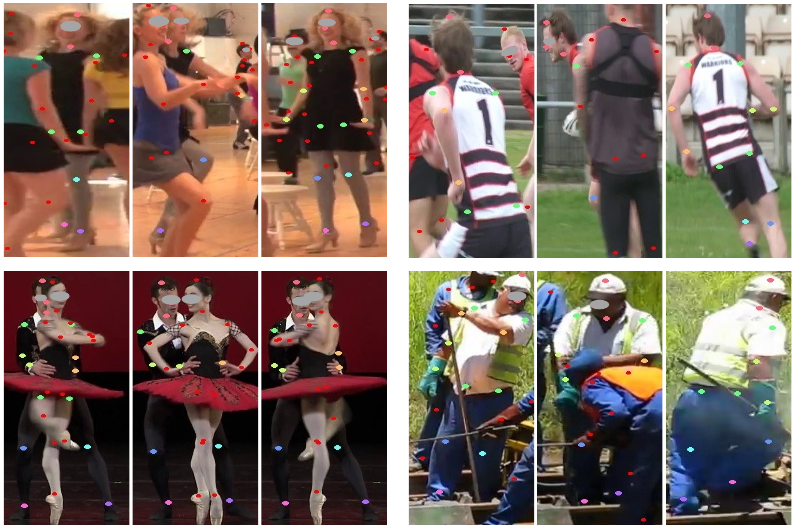}
    \caption{{\OccludedPT}}
    \label{fig:dataset}
  \end{subfigure}
  \caption{
    (a) Our proposed Batch-wise Inter-Person Occlusion (BIPO) data augmentation creates artificial person occlusions that are consistent across image, prompt, and human parsing labels. BIPO is crucial to enforce the model to rely on the input prompts. (b) Four identities with their keypoints labels from our {\OccPT} dataset.
    }
\end{figure}

\textbf{Design motivations}: Given that a triplet loss with batch-hard mining \cite{triplet} is employed during training, using images from the same training batch to generate occluders yields hard positive and negative samples for mining triplets, thereby creating challenging scenarios for contrastive learning. BIPO is illustrated in \cref{fig:bipo} and its significance for KPR is further discussed in \cref{section:pipeline_intuition}.


\subsection{Inference} \label{section:testing}
At inference, {\model} processes each input image together with their respective keypoints prompts, and produces $K$ part-based representations \{$f_1$, ..., $f_K$\} of the {\reid} target and their binary visibility scores \{$v_1$, ..., $v_K$\}.
Similar to previous part-based works \cite{VGTri, ISP}, the global distance $dist_{\rm{global}}^{qg}$ between a query image $q$ and a gallery image $g$ is computed as the average cosine distance $dist_{\rm{cos}}$ of the part-based embeddings $f_i$ that are visible in both samples:

{\small{
\begin{equation} \label{eq:test_dist}
dist_{\rm{global}}^{qg} = 
\frac{  \sum\limits_{i \in \{1, ..., K\}} \Big(v^{q}_{i} \cdot v^{g}_{i} \cdot dist_{\rm{cos}}(f_{i}^{q}, f_{i}^{g}) \Big) }
{  \sum\limits_{i \in \{1, ..., K\}} \big(v^{q}_{i} \cdot v^{g}_{i} \big) }\ .
\end{equation}
}}

\subsection{KPR Pipeline Motivation and Intuition} \label{section:pipeline_intuition}

Our pipeline comprises two closely related architectural components.
First, the prompt tokenizer, which encodes semantic keypoints information to disambiguate the ReID target, thereby providing prior information about body part location to drive feature extraction.
Second, the Part-based Head (PBH), responsible for localizing the body parts of the target and constructing its part-based features.

While the Part-based Head is explicitly supervised with human parsing labels at training to learn human topology, there is no specific loss incentivizing the model to leverage information from the input prompt.
Additionally, multi-person occlusions are rare in most ReID training sets, giving the model little incentive to rely on these prompts to localize the target person.
For this reason, just adding a prompt tokenizer and training the pipeline as is produces a KPR model with weak reliance on the input prompts, and therefore low performance in multi-person occlusion scenarios.

To address these issues, our proposed BIPO data augmentation plays a critical role.
By generating artificial multi-person occlusions during training, the model is compelled to rely more on the prompts, as they represent the sole reliable source of information to distinguish the true ReID target from the occluder.

Consequently, incorporating BIPO during training is crucial to produce a model that is proficient at leveraging prompt information and effectively handling multi-person occlusions. 
Finally, designing the Part-based Head without explicit reliance on the prompt offers an additional advantage: it renders our model \textit{prompt-optional}, enhancing robustness against noisy or missing prompts while still enabling effective localization of the target's body parts.

\section{Experiments} \label{section:experiments}

                                        
\subsection{Evaluation Benchmarks and Metrics} \label{section:datasets}

We assess our model's performance on four types of \textit{person retrieval} setups: non-occluded (i.e., holistic) on \textbf{Market-1501} \cite{market1501}, object occlusions on \textbf{Occluded-ReID} \cite{Zhuo2018} and \textbf{Partial-ReID}\footnote{We use the occluded (non-cropped) images as queries.} \cite{partial_reid}, multi-person occlusions on our below introduced \textbf{{\OccludedPT}} dataset, and finally both objects and multi-persons occlusions on \textbf{Occluded-Duke} \cite{PGFA}.
Two standard person retrieval metrics are reported: the CMC at \textit{Rank-1} and the \textit{mAP}. 
Performances are evaluated in a single query setting and without re-ranking \cite{re-rank}.
Furthermore, we evaluate our model on \textit{multi-person pose tracking} with the \textbf{PoseTrack21} \cite{Doering2022} dataset.
We report the \textit{HOTA} \cite{Luiten2021} performance metric, and relevant sub-metrics such as association accuracy (\textit{AssA}), detection accuracy (\textit{DetA}), and the number of identity switches (\textit{IDs}) \cite{Leal-Taixe2015}.

\subsection{Proposed {\OccludedPT} Dataset and Annotations} \label{section:occ_pt}

We introduce {\OccludedPT} (or simply {\OccPT}), a new {\reid} dataset we built out of the annotation available with PoseTrack21 \cite{Doering2022}, a popular video benchmark for multi-person pose tracking, that features keypoints and cross-video identity annotations.
Unlike previous {\reid} datasets focused on street surveillance, {\OccPT} consists of images from everyday life videos, primarily from sports activities, as illustrated in \cref{fig:dataset}.
{\OccPT} is divided into a train/test that includes 1000/1411 identities with 17.898/13.412 images from 474/170 videos, which is roughly equivalent in terms of scale to other popular {\reid} datasets \cite{market1501, PGFA} (\cref{tab:reid_datasets} of the supp. materials). 
To assess the {\reid} model's performance in multi-person occlusion scenarios, we select the most cluttered images of each identity in the test set as query samples, and the remaining test images as gallery samples.
Cluttered images corresponds to multi-persons occlusions scenarios where either the front (occluding) or back (occluded) person is the {\reid} target, to evaluate the model ability to re-identify individuals in both scenarios.
We provide further details about our proposed dataset in the supp. materials.
{\OccPT} is challenging as persons within the same video exhibit a high degree of visual resemblance since they often wear similar sports kits.

\textbf{Keypoint Annotations for Standard {\reid} Datasets}: Since conventional {\reid} datasets \cite{market1501, PGFA, Zhuo2018} do not contain keypoint annotations, we generate pseudo-labels for them with the PifPaf \cite{pifpaf} pose estimator.
These pseudo-labels include keypoint annotations in COCO \cite{coco} format for each image, as illustrated in \cref{fig:bipo}. 
When multiple skeletons are detected in an image, we make the assumption the one with its head closer to the top center part of the image is the intended target, 
and mark it with a ``\textit{is\_target}" attribute.
Other skeletons are used as negatives.
More details are provided in the supplementary materials.

\subsection{Implementation Details} \label{section:implementation}
We evaluate our model with both the ImageNet \cite{imagenet} (\textbf{{\model}\textsubscript{IN}}) and the SOLIDER \cite{SOLIDER} (\textbf{{\model}\textsubscript{SOL}}) pre-trained Swin backbones.
SOLIDER is a human-centric foundation model trained on the large-scale LUPerson \cite{LUPerson} dataset in a self-supervised way.
Following \cite{bpbreid}, the number of body parts $K$ is empirically set to $8$ for Occluded-Duke/{\OccPT}, and $5$ otherwise.
The training procedure is mainly adopted from TransReID \cite{transreid} and SOLIDER \cite{SOLIDER}.
We apply our \textit{BIPO} data augmentation (\cref{section:bipo}) with a $0.3$ probability.
The model is trained for 120 epochs with a batch size of 64 and a cosine annealing \textit{lr} scheduler.
For {\model}\textsubscript{IN}/{\model}\textsubscript{SOL}, images are resized to a width of $128$ and a height of $256$/$384$.
The Gaussian kernel standard deviation $\alpha_{G}$ (\cref{section:tokenization}) is set to $0.1$.



\subsection{Comparison with State-of-the-Art Methods} \label{section:sota}

In \Cref{table:sota_all_datasets_small}, we present a comparative analysis of {\model}, when \textbf{using prompts} (\textit{KPR}) and \textbf{not using prompts}\footnote{The prompt tokenizer is complitely removed at both training and inference.} (\textit{KPR w/o prompt}), against all leading occluded {\reid} methods.
Our method emerges as the top performer overall. 

\textbf{{\OccPT} and Occ-Duke:}
Unlike the other three datasets, both {\OccPT} and Occluded-Duke contain a substantial number of images with multi-person occlusions, making them suitable environments to assess the effectiveness of our promptable method.
On these two challenging occluded datasets, our proposed model {\model} outperforms all previous occluded \cite{bpbreid, FED, OPReID} and pose-guided \cite{PFD, VGTri, HOReID} methods.

\begin{table}[t!]
\centering
\caption{
Comparison of {\model} with SOTA methods.
Results in \textit{Italic} are not provided in the original paper but reproduced by ourselves.
The 1$^{st}$/2$^{nd}$/3$^{rd}$ best scores are indicated with $\textcolor{red}{^{1/2/3}}$.
}
\label{table:sota_all_datasets_small}
\resizebox{0.8\columnwidth}{!}{
\begin{tabular}{|c|cc|cc|cc|cc|cc|}
\hline
\multicolumn{1}{|c|}{\multirow{2}{*}{\makecell[c]{Datasets}}} &
\multicolumn{2}{c|}{\multirow{2}{*}{\makecell[c]{Market-\\1501}}} &
\multicolumn{2}{c|}{\multirow{2}{*}{\makecell[c]{Occluded-\\reID}}} &
\multicolumn{2}{c|}{\multirow{2}{*}{\makecell[c]{Partial-\\reID}}} &
\multicolumn{2}{c|}{\multirow{2}{*}{\makecell[c]{\textbf{Occluded}-\\\textbf{PoseTrack}}}} &
\multicolumn{2}{c|}{\multirow{2}{*}{\makecell[c]{Occluded-\\Duke}}}
\\
 & & & & & & & & & & \\
 \hline
Type &
\multicolumn{2}{c|}{\multirow{1}{*}{\makecell[c]{Holistic}}} &
\multicolumn{8}{c|}{\multirow{1}{*}{\makecell[c]{Occluded}}} \\
\hline
Object Occlusions &
\multicolumn{2}{c|}{} &
\multicolumn{2}{c|}{$\checkmark$} &
\multicolumn{2}{c|}{$\checkmark$} &
\multicolumn{2}{c|}{} &
\multicolumn{2}{c|}{$\checkmark$}
\\
\hline
Person Occlusions &
\multicolumn{2}{c|}{} &
\multicolumn{2}{c|}{} &
\multicolumn{2}{c|}{} &
\multicolumn{2}{c|}{$\checkmark$} &
\multicolumn{2}{c|}{$\checkmark$}
\\
\hline
\hline
Methods & {R-1}&{mAP}&{R-1}&{mAP}&{R-1}&{R-3}&{R-1}&{mAP}&{R-1}&{mAP} \\
\hline
BoT \cite{BoT} &          94.5 & 85.9 & 58.4 & 52.3 & - & - & \textit{78.8} & \textit{69.7} & 51.4 & 44.7 \\
PCB \cite{PCB} &         93.8 & 81.6 & - & - & - & - & \textit{81.7} & \textit{71.2} & 51.2 & 40.8 \\  
VGTri \cite{VGTri}&     -    & -    & 81.0 & 71.0 & 85.7 & 93.7$\textcolor{red}{^{3}}$ & - & - & 62.2 & 46.3 \\
PVPM \cite{PVPM} &       -    & -    & 66.8 & 59.5 & 78.3 & - & - & - & -    & -    \\ 
HOReID \cite{HOReID} &   94.2 & 84.9 & 80.3 & 70.2 & 85.3 & 91.0 & - & - & 55.1 & 43.8 \\
ISP \cite{ISP} &         95.3 & 88.6 & - & - & - & - & - & - & 62.8 & 52.3 \\
PAT \cite{PAT} &         95.4 & 88.0 & 81.6 & 72.1 & 88.0$\textcolor{red}{^{3}}$ & 92.3 & - & - & 64.5 & 53.6 \\  
TRANS \cite{transreid} & 95.2 & 88.9 & - & - & - & - & \textit{83.5} & \textit{73.4} & 66.4 & 59.2 \\  %
SOLIDER \cite{SOLIDER} & \textbf{96.9}$\textcolor{red}{^{1}}$ & \textbf{93.9}$\textcolor{red}{^{1}}$ & - & - & - & - & \textit{84.4} & \textit{76.1}$\textcolor{red}{^{3}}$ & \textit{71.2} & \textit{61.9} \\  
SSGR \cite{OPReID} &      96.1 & 89.3 & 78.5 & 72.9 & - & - & - & - & 69.0 & 57.2 \\
FED \cite{FED} &          95.0 & 86.3 & \textbf{86.3}$\textcolor{red}{^{1}}$ & 79.3$\textcolor{red}{^{3}}$ & 84.6 & - & - & - & 68.1 & 56.4 \\
BPBreid \cite{bpbreid} & 95.7 & 89.4 & 82.9 & 75.2 & - & - & \textit{84.9} & \textit{75.5} & 75.1 & 62.5 \\
PFD \cite{PFD} &         95.5 & 89.7 & 83.0 & 81.5$\textcolor{red}{^{2}}$ & - & - & - & - & 69.5 & 61.8 \\
\hline
{\model}\textsubscript{IN} w/o prompt & 95.6 & 88.7 & 83.3 & 78.2 & 81.7 & 86.0 & 85.3 & 75.4 & 75.8 & 63.4 \\
\rowcolor{Pink}
{\model}\textsubscript{IN} & 95.9 & 89.6 & 85.4$\textcolor{red}{^{2 }}$ & 79.1 & 86.0 & 90.0 & \textbf{92.3}$\textcolor{red}{^{1}}$ & \textbf{82.3}$\textcolor{red}{^{1}}$ & 79.8$\textcolor{red}{^{3}}$ & 67.1$\textcolor{red}{^{3}}$ \\
\hline
{\model}\textsubscript{SOL} w/o prompt & 96.6$\textcolor{red}{^{3}}$ & 93.0$\textcolor{red}{^{3}}$ & 80.0 & 78.5 & 90.3$\textcolor{red}{^{2}}$ & 93.7$\textcolor{red}{^{2}}$ & 86.1$\textcolor{red}{^{3}}$ & 75.8 & 82.5$\textcolor{red}{^{2}}$ & 73.3$\textcolor{red}{^{2}}$ \\
\rowcolor{Pink}
{\model}\textsubscript{SOL} & 96.6$\textcolor{red}{^{2}}$ & 93.2$\textcolor{red}{^{2}}$ & 84.8$\textcolor{red}{^{3}}$ & \textbf{82.6}$\textcolor{red}{^{1}}$ & \textbf{90.7}$\textcolor{red}{^{1}}$ & \textbf{94.0}$\textcolor{red}{^{1}}$ & 90.6$\textcolor{red}{^{2}}$ & 81.2$\textcolor{red}{^{2}}$ & \textbf{84.3}$\textcolor{red}{^{1}}$ & \textbf{75.1}$\textcolor{red}{^{1}}$ \\

\hline
\end{tabular}
}
\end{table}

Our \textit{keypoint promptable} approach offers several key advantages over the previous \textit{pose-guided} methods, including: 1) the ability to handle multi-person occlusions; 2) the capability to process both positive and negative keypoints; 3) a \textit{prompt-aware appearance encoding}, leading to better feature disentanglement between multiple persons; and 4) the flexibility of prompts being optional.
We refer readers to \cref{section:related_work} for a detailed discussion about these key differences and a comparison to previous methods.
On our proposed {\OccPT}, we can see that the best-performing method is the foundation model SOLIDER, but it still lies about 6\% behind our {\model} solution.
Even BPBreID \cite{bpbreid}, which is specialized for occluded scenarios, demonstrates far lower performance, since it is not designed to address the Multi-Person Ambiguity.
Our experiments show that integrating keypoint prompts boosts performance by at least 7.0\% Rank-1 and 6.2\% mAP, since it helps in mitigating the negative impact of MPA.
On Occluded-Duke, the most popular occluded dataset, we outperform previous SOTA methods by at least +12.6\% mAP and +9.2\% rank-1, significantly narrowing the performance gap between holistic and occluded {\reid}.

\textbf{Market-1501, Occluded-ReID and Partial-reID:}
These three datasets primarily feature single individuals in their images, which means multi-person ambiguities are rare, and therefore prompts are not exploited to their full potential. 
On Market-1501, our method performs on par with the state-of-the-art.
However, performance levels have plateaued in recent years on this dataset.
In contrast, on Occluded-ReID and Partial-reID, using a prompt noticeably improves performance by helping distinguish the individual features from occluding objects.
Additionally, using prompt is beneficial when facing this cross-domain\footnote{Occ-ReID and Partial-ReID have no train set, so Market-1501 is used for training.} setting.
These findings are consistent with earlier studies \cite{PFD, HOReID, PVPM, FED} that also leveraged pose estimation to attain strong performance on this dataset.


\noindent 
\begin{figure}[t]
\begin{minipage}{0.5\textwidth}
    \captionof{table}{Multi-person pose tracking performance in videos on PoseTrack21 \cite{Doering2022}.}
    \label{table:tracking}
    \centering
    \resizebox{\textwidth}{!}{
    \begin{tabular}{|c|c|c|c|c|}
    \hline
    Method & HOTA$\uparrow$ & DetA$\uparrow$ & AssA$\uparrow$ & IDs$\downarrow$ \\ 
    \hline
    TRMOT \cite{TRMOT} & 46.8 & 40.9 & 55.0 & - \\
    FairMOT \cite{FairMOT} & 53.5 & 47.4 & 61.4 & - \\
    Tracktor++ \cite{Tracktor} & 58.3 & 52v.7 & 65.4 & - \\
    CorrTrack + ReID \cite{Doering2022} & 56.9 & 51.3 & 64.2 & - \\
    \hline
    {\model}Track w/o prompt & 61.5 & 58.4 & 65.8 & 3830 \\  
    \hline
    \rowcolor{Pink}
     & \textbf{63.0} & \textbf{59.0} & \textbf{68.3} & \textbf{3344} \\  
     \rowcolor{Pink}
    \multirow{-2}{*}{\makecell[c]{{\model}Track}} & \textcolor{PositiveGreen}{(+1.5)} & \textcolor{PositiveGreen}{(+0.6)} & \textcolor{PositiveGreen}{(+2.5)} & \textcolor{PositiveGreen}{(-486)} \\
    \hline
    \end{tabular}}
\end{minipage}%
\hfill
\begin{minipage}{0.45\textwidth}
    \centering
    \includegraphics[width=\linewidth]{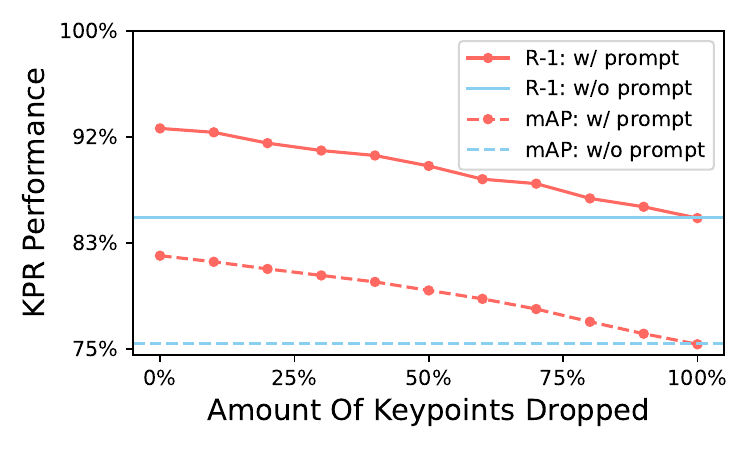}
    \captionof{figure}{Prompt optionality study}
    \label{fig:optional_prompt_plot}
\end{minipage}
\end{figure}

\subsection{Multi-Person Pose Tracking in Videos} \label{section:tracking}

We introduce a simple yet robust pose tracker, referred to as \textit{{\model}Track}, built upon our promptable {\reid} method.
{\model}Track combines the YOLOX \cite{YOLOX} object detector with the HRNet \cite{hrnet} pose estimator.
For each detected person, it extracts part-based {\reid} features utilizing their keypoints as positive prompts and keypoints from other individuals \textit{within the detected person bbox} as negative prompts.
Detections are matched in an online fashion with the Hungarian algorithm based on the distance function introduced in \cref{eq:test_dist}.
Only tracklet-detection pairs with a distance below a threshold of $0.2$ are considered for matching.
Therefore, our proposed tracker relies solely on appearance information, meaning spatio-temporal cues are omitted. 
Pose tracking results are reported in \cref{table:tracking}.
Despite relying on ReID only, {\model}Track achieves state-of-the-art performance, demonstrating that an occlusion-robust {\reid} model has strong tracking capabilities.
Finally, our two last experiments in \cref{table:tracking} reveal that employing keypoint prompts significantly enhances tracking performance.
This improvement can be attributed to the enhanced communication between the detector and the ReID via the keypoint prompts.
This leads to improved association accuracy and reduced identity switches, especially when multiple targets intersect paths (i.e., multi-person occlusion scenarios).
Further information regarding our baseline tracker is provided in the supplementary materials.


\subsection{Ablation Studies} \label{section:ablation_studies}
In this section, we analyze the performance impact of our architectural choices.

\subsubsection{Components of {\model}} \label{section:components_ablation}


\begin{table}[t!]
\centering
\caption{
Ablation study of our proposed KPR architecture. 
PBH stands for ``Part-based Head", meaning we use multiple part-based embeddings with visibility scores instead of a single global embedding (\cref{section:pbh}).
MSF stands for the ``Multi-Scale Features" module added to Swin (\cref{section:encoding}). 
Prompt $\oplus$/$\ominus$ refers to positive/negative prompts (\cref{section:tokenization}).
{\randocc} stands for the ``Batch-wise Inter-Person Occlusion" data augmentation (\cref{section:bipo}). 
SOL stands for the pre-trained weights from SOLIDER \cite{SOLIDER} (\cref{section:implementation}).
}
\label{table:main_ablation_study}
\resizebox{0.85\columnwidth}{!}{
\begin{tabular}{|c|c|c|c|c|c|c|c|c|c|c|}
\hline
\multirow{3}{*}{Id} &\multicolumn{6}{c|}{Main components of {\model}} & \multicolumn{4}{c|}{Dataset} \\ 
\cline{2-11} 
 & \multirow{2}{*}{PBH} & \multirow{2}{*}{MSF} &  \multicolumn{2}{c|}{Prompt} & \multirow{2}{*}{BIPO} & \multirow{2}{*}{SOL} & \multicolumn{2}{c|}{Occluded-PoseTrack} & \multicolumn{2}{c|}{Occluded-Duke} \\
\cline{4-5} \cline{8-11} 
  &  &  & \multicolumn{1}{c|}{$\oplus$} & \multicolumn{1}{c|}{$\ominus$} &  &  & \multicolumn{1}{c|}{R-1} & \multicolumn{1}{c|}{mAP} & \multicolumn{1}{c|}{R-1} & \multicolumn{1}{c|}{mAP} \\ 
\hline
1 &  &  &  &  &  &                                                                  & 84.8                           & 74.7                           & 61.8                       & 52.4 \\  %
2 & \checkmark &  &  &  &  &                                                        & 83.3 (\textcolor{gray}{-1.5})  & 72.2 (\textcolor{gray}{-2.5})  & 67.6 (\textcolor{PositiveGreen}{+5.8}) & 54.2 (\textcolor{PositiveGreen}{+1.8}) \\  %
3 & \checkmark & \checkmark &  &  &  &                                              & 84.8 (\textcolor{PositiveGreen}{+0.0})  & 76.1 (\textcolor{PositiveGreen}{+1.4}) & 68.9 (\textcolor{PositiveGreen}{+7.1}) & 56.8 (\textcolor{PositiveGreen}{+4.4}) \\  %
4 & \checkmark & \checkmark & \checkmark &  &  &                                    & 88.8 (\textcolor{PositiveGreen}{+4.0}) & 80.4 (\textcolor{PositiveGreen}{+5.7}) & 70.8 (\textcolor{PositiveGreen}{+9.0}) & 58.7 (\textcolor{PositiveGreen}{+6.3}) \\  %
5 & \checkmark & \checkmark & \checkmark & \checkmark &  &                          & 90.0 (\textcolor{PositiveGreen}{+5.2}) & 80.7 (\textcolor{PositiveGreen}{+6.0}) & 73.8 (\textcolor{PositiveGreen}{+12.0}) & 62.0 (\textcolor{PositiveGreen}{+9.6}) \\  %
6 &  &  & \checkmark & \checkmark &  &                          & 87.2 (\textcolor{PositiveGreen}{+2.4}) & 78.9 (\textcolor{PositiveGreen}{+4.2}) & 69.0 (\textcolor{PositiveGreen}{+7.2}) & 59.7 (\textcolor{PositiveGreen}{+7.3}) \\  %
7 & \checkmark & \checkmark &  &  & \checkmark &                                    & 86.1 (\textcolor{PositiveGreen}{+1.3}) & 76.7 (\textcolor{PositiveGreen}{+2.0}) & 75.8 (\textcolor{PositiveGreen}{+14.0}) & 63.4 (\textcolor{PositiveGreen}{+11.0}) \\  %
\rowcolor{Pink}
8 & \checkmark & \checkmark & \checkmark & \checkmark & \checkmark &                & \textbf{92.3} (\textcolor{PositiveGreen}{+6.5}) & \textbf{82.3} (\textcolor{PositiveGreen}{+7.6}) & 79.8 (\textcolor{PositiveGreen}{+18.0}) & 67.1 (\textcolor{PositiveGreen}{+14.7}) \\  %
\rowcolor{Pink}
9 & \checkmark & \checkmark & \checkmark & \checkmark & \checkmark & \checkmark     & 90.6 (\textcolor{PositiveGreen}{+5.8}) & 81.2 (\textcolor{PositiveGreen}{+6.5}) & \textbf{84.3} (\textcolor{PositiveGreen}{+22.5}) & \textbf{75.1} (\textcolor{PositiveGreen}{+22.7}) \\  %
\hline
\end{tabular}
}
\end{table}

As demonstrated in \Cref{table:main_ablation_study}, all our proposed modules introduced in \cref{section:methodology} play a crucial role in improving the overall {\reid} performance. 
Exp. 2 demonstrates the importance of the Part-based Head to compare only mutually visible body parts when facing occluded persons, compared to using a single global feature (as in Exp. 1). 
Exp. 3 is consistent with claims from previous works \cite{BoT, ISP} that increasing the feature map resolution is crucial to achieve fine-grained re-identification and better overall performance.
Exp. 4 and 5 illustrate the benefits of positive and negative prompts to disambiguate the intended {\reid} target from other persons.
Exp. 6 show the importance of 
Exp. 7 demonstrates the importance of our {\randocc} augmentation to generate artificial occlusions at training, since most occluded {\reid} benchmarks feature mainly occluded samples in their test set and not in their training set. 
Exp. 8, in comparison with Exp. 5 and 7, demonstrates how generating these artificial multi-person occlusions at training is crucial to teach the model to rely on the input prompt to identify the intended {\reid} target among all candidates.
We highlight the advantage of using the SOLIDER foundation model as pre-trained weights in Exp. 9. 
However, SOLIDER does not enhance performance on the {\OccPT} dataset, likely because of a domain gap: SOLIDER focuses on street surveillance, while {\OccPT} predominantly features sports images from handheld cameras.




\subsubsection{A Prompt-Optional Method} \label{section:impact_occlusions}

In \Cref{fig:optional_prompt_plot}, we analyze the test-time robustness of {\model} against noisy or missing prompts on {\OccPT}, by increasingly removing a higher percentage of keypoints from the input prompt.
The red curve represents the performance of our proposed promptable {\model} model, while the blue constant line illustrates the performance of the non-promptable version, which is unaffected by the proportion of keypoints removed from the input prompt.
As illustrated, {\model} significantly outperforms the non-promptable baseline with a full prompt and remains competitive with few or no keypoints, highlighting its prompt-optional capability.
We provide an additional study of the performance w.r.t. the amount of multi-person occlusion in the supplementary materials.

\subsubsection{Qualitative Assessment}
\begin{figure}[t!]
    \centering
    \begin{subfigure}[b]{1.\textwidth} 
        \centering
        \includegraphics[height=2.7cm,width=\textwidth]{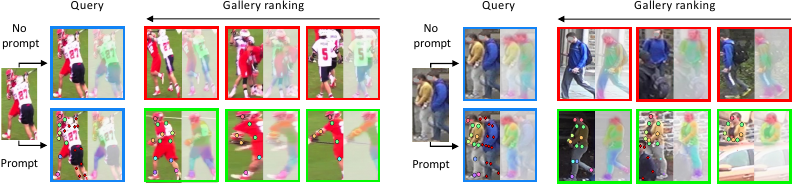}
        \caption{
        Ranked gallery samples for a given query, when the input prompt is disabled/enabled.
        Green/red borders are correct/incorrect matches.The model focuses on the wrong target without the prompt.
        }
        \label{fig:qualitative_results_1}
    \end{subfigure}
    \\
    \begin{subfigure}[b]{0.27\textwidth} 
        \centering
        \includegraphics[height=2.7cm,keepaspectratio]{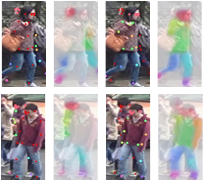}
        \caption{Part-attention maps for various input prompts.}
        \label{fig:qualitative_results_2}
    \end{subfigure}
    \hfill 
    \begin{subfigure}[b]{0.7\textwidth} 
        \centering
        \includegraphics[height=2.7cm,keepaspectratio]{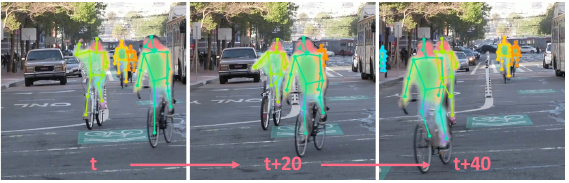}
        \caption{
        Pose tracking results: all track IDs, depicted by unique skeleton colors, are maintained after a long occlusion by the front cyclist.
        }
        \label{fig:qualitative_results_3}
    \end{subfigure}
    
    \caption{
        Qualitative results on multi-person pose tracking and person retrieval.
        Red/pastel dots indicate negative/positive prompts.
        Colored heatmaps illustrates the part-attention maps of {\model} described in \cref{section:pbh}, with one color per body part.
    }
    \label{fig:qualitative_results}
\end{figure}

\Cref{fig:qualitative_results} demonstrates the impact of keypoint prompting on the part-attention maps, on person retrieval, and on pose tracking.
As illustrated, {\model} effectively leverages the input prompts to localize the intended target, and is therefore robust to occlusions and multi-person ambiguities.

\section{Conclusion}

In this work we propose {\model}, the first promptable {\reid} model, designed to address occluded scenarios and multi-person ambiguities arising from bounding-box-based inputs.
Our model offers practical flexibility by being prompt-optional, achieving state-of-the-art performance without prompts, and outperforming this baseline with prompts. 
Furthermore, our model outperforms the state-of-the-art on three popular {\reid} benchmarks and on our novel {\OccPT} dataset.
We finally demonstrate KPR's potential for pose tracking in videos.
Our codebase, keypoint labels and proposed dataset will be released to encourage further research on this new promptable {\reid} paradigm.


\textbf{Acknowledgment:}
This work has been funded by Sportradar, by the Walloon region project ReconnAIssance, and by the FNRS.




\bibliographystyle{splncs04}

\bibliography{
bib/abbreviation-short,
bib/eccv24
}

\clearpage
\appendix
\pagebreak
\section*{Keypoint Promptable Re-Identification: Supplementary Materials}

In the upcoming sections, we present additional details regarding the novel {\OccludedPT} dataset and our {\model}Track pose tracking method.
Furthermore, we offer supplementary qualitative and quantitative results to provide a more comprehensive perspective on our research.
Our code, annotations, and proposed dataset are available at \url{https://github.com/VlSomers/keypoint_promptable_reidentification}.
Our codebase was forked from the \textbf{Torchreid}\footnote{\url{https://github.com/KaiyangZhou/deep-person-reid}} framework and BPBreID\footnote{\url{https://github.com/VlSomers/bpbreid}} \cite{bpbreid}.
Pose-tracking results were obtained using the Tracklab framework for multi-object tracking \cite{Joos2024Tracklab}.

\subsection*{Study on the Robustness to Multi-Person Occlusions} \label{section:impact_occlusions}

\begin{figure}
  \center
  \includegraphics[width=0.45\linewidth]{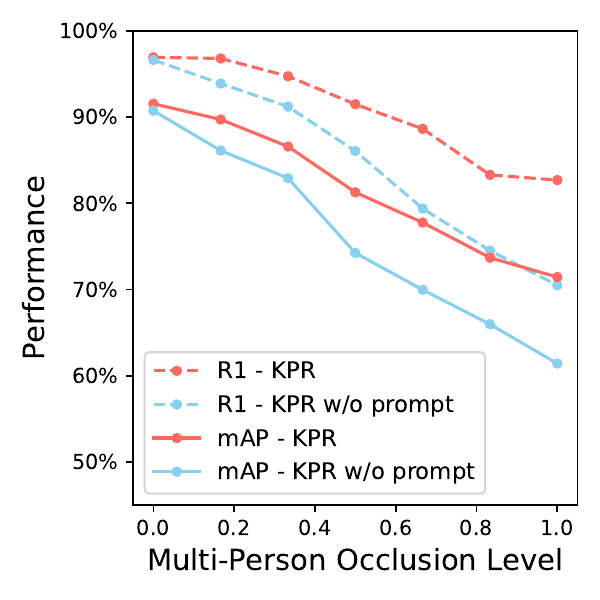}
  \caption{
    Performance w.r.t. the multi-person occlusion level of the queries.
  }\label{fig:occlusions_plot_1}
\end{figure}

In \Cref{fig:occlusions_plot_1}, we present a performance comparison between {\model} and its non-promptable version under various occlusion levels on {\OccPT}.
To quantify the Multi-Person Occlusion Level (MPOL) of a query sample $i$ in the query set $Q$, we compute the difference between the number of negative keypoints $N_i$ and the number of positive keypoints $P_i$, and then normalize this value over the set $Q$ to obtain a percentage as follows:

{\small{
\begin{equation} \label{eq:occ_level}
\rm{MPOL}_{i} = \frac{(N_{i}-P_{i}) - \underset{j \in Q}{\mathrm{min}}(N_{j}-P_{j})}{ \underset{j \in Q}{\mathrm{max}}(N_{j}-P_{j}) - \underset{j \in Q}{\mathrm{min}}(N_{j}-P_{j}) }.
\end{equation}
}}

As shown in the plots, the utilization of prompts leads to at least 10\% performance enhancement in scenarios with heavy multi-person occlusions. 
Consequently, these experiments demonstrate the efficacy of keypoints prompts to disambiguate the intended {\reid} target among multiple pedestrians.

Furthermore, keypoint prompts contribute to better part-based feature extraction, which explains the performance boost in scenarios with low occlusion scores.

\subsection*{{\OccludedPT} Dataset Details} \label{section:dataset}

As introduced in \Cref{section:dataset}, {\OccludedPT} (or simply {\OccPT}), is a new {\reid} dataset we built out of the annotation available with PoseTrack21 \cite{Doering2022}.
PoseTrack21 is a popular video benchmark for multi-person pose tracking, that features keypoints and cross-video identity annotations.
We provide a train/test split of {\OccludedPT} that is based on the original train/validation split of PoseTrack21.
Important numbers about each split are summarized in Table \ref{tab:posetrack_reid_dataset}.
Similar to the original dataset, the train and test splits do not have overlapping video scenes and therefore do not share identities.
In order to build the {\OccPT} dataset, we randomly sample 1000 identities from the PoseTrack21 train set and keep all 1411 identities from the test set, and uniformly sample each tracklet, so that each identity has at least 4 image crops and maximum 20.
Finally, we divide the test set into a gallery set and a query set, so that 20\% of each identity images are added to the query set and the remaining 80\% samples are added to the gallery set.
Similar to existing dataset specialized on occlusions \cite{PGFA}, we choose the 20\% most occluded samples from each identity as query samples.
The occlusion score of each sample is computed using \cref{eq:occ_level}.
This sampling strategy renders the dataset more relevant to evaluate methods that are good at addressing occlusions and multi-person ambiguity.
Compared to previous datasets that only compare query samples against gallery samples captured from a different camera viewpoint, we compare each query sample against all galleries, whether they are from a different video or not.
Compared to previous sport re-identification datasets \cite{soccernet22, deepsportradarv1}, PoseTrack-ReID contains identity labels that are valid across the entire dataset, since a given identity can be spotted in multiple videos.
We provide numbers to compare the scale of our dataset with existing popular {\reid} benchmarks in \cref{tab:reid_datasets}.
We encourage researchers to not use re-ranking \cite{re-rank} to evaluate their {\reid} method on this dataset.

\begin{table}[h!]
\centering
\begin{tabular}{c|ccc}
Subset & \# Ids & \# Imgs & \# Videos \\ \hline
train & 1000 & 17898 & 474 \\
query & 1379 & 2581 & 163 \\
gallery & 1411 & 10831 & 170 \\ \hline


\end{tabular}%
\caption{Characteristics of each PoseTrack-ReID subset.}
\label{tab:posetrack_reid_dataset}
\end{table}

\begin{table*}[t]
\centering
\resizebox{\columnwidth}{!}{%
\begin{tabular}{l|ccccccc}
Dataset & Type & \# Ids & \# Cams & \# Imgs & Release & Train set & Occlusions\\ \hline
Market-1501 \cite{market1501} & Street & 1501 & 6 & 32,217 & 2015 & \checkmark &  \\
DukeMTMC-ReID \cite{duke-mtmc} & Street & 1812 & 8 & 36,441 & 2017 & \checkmark &  \\
MSMT17 \cite{Wei2017} & Surveillance & 4101 & 15 & 126,441 & 2018 & \checkmark &  \\
Occluded-Duke \cite{PGFA} & Street & 1812 & 8 & 35,489 & 2019 & \checkmark & \checkmark  \\
Occluded-ReID \cite{Zhuo2018} & Outdoor & 200 & - & 2,000 & 2018 &  & \checkmark \\
SoccerNet-ReID \cite{soccernet22} & Soccer & 243,432 & - & 340,993 & 2022 & \checkmark &  \\
DeepSportradar-ReID \cite{deepsportradarv1} & Basketball & 486 & - & 9,529 & 2022 & \checkmark &  \\
\textbf{{\OccPT} (ours)} & Multi-sport & 2411 & - & 31310 & 2024 & \checkmark & \checkmark \\ 
\hline
\end{tabular}%
}
\caption{Comparing our proposed {\OccludedPT} with other popular {\reid} datasets.}
\label{tab:reid_datasets}
\end{table*}


\subsection*{Keypoint and Human Parsing Annotations}

Our method requires two types of annotations: the keypoints labels that are used for prompting at both train and test time, and the human parsing labels that are only used at training time by the part-prediction loss.

For Market-1501, Occluded-Reid, Partial-ReID and Occluded-Duke, we generate keypoint labels with the PifPaf \cite{pifpaf} pose estimation model as described in \cref{section:occ_pt}, and employ the human parsing labels provided by BPBreID \cite{bpbreid}.
For Occluded-Duke only, we manually fixed $+/- 10\%$ of the query samples that either had a person missed by PifPaf or the wrong person labelled as the target, following our heuristic introduced in \cref{section:occ_pt}.
This heuristic stated that the person with its head closer to the top center part of the image was chosen as the target person when multiple skeletons were detected.
Remaining gallery and training samples were not fixed manually.

For our proposed {\OccPT} dataset, we simply employ the keypoint labels already available in the original PoseTrack21 dataset.
To generate the human parsing labels, we employ the same methodology as the one described in \cite{bpbreid}, with a small difference regarding the generation of the segmentation mask of the {\reid} target.
This segmentation mask is required to filter out the Pif and Paf confidence fields activations that belongs to other non-target persons.
Different from BPBreID's authors who employed a MaskRCNN segmentation model, we employ Segment Anything \cite{SAM} and prompt it with the target's person keypoints to generate a consistent segmentation mask.
We also tried to employ SAM to directly generate the human parsing labels without PifPaf, but we found out SAM was bad at segmenting body parts, even when prompted with accurate keypoints.
Combining SAM (to segment the target person) and PifPaf (to assign a body part to each pixel in the bounding box image) led to the most accurate human parsing labels.
We refer readers to \cite{bpbreid} for more details on the human parsing labels generation methodology used for existing ReID datasets.


\subsection*{{\model}Track Details}
In this Section, we provide further details on our proposed pose-tracking method {\model}Track.
{\model}Track employs two ByteTrack's components: 1) the tracklet life-cycle management mechanism and 2), the multi-stage association strategy based on the confidence level of the detections in the current frame.
More details can be found in the original ByteTrack's paper \cite{Zhang2021ByteTrack}.

To obtain strong person detection performance, we fine-tuned the YOLOX objector detector and the HrNet pose estimator on the PoseTrack21 train set.
As stated in \Cref{section:tracking}, our proposed tracker does not employ spatio-temporal information, as opposed to almost all existing state-of-the-art multi-object tracker.
However, our tracker can be easily extended with additional spatio-temporal cues (e.g. a Kalman Filter with bounding box IOU) to further improve performance.

As a re-identification method providing appearance cues to the tracking pipeline, we employ our proposed {\model} method trained on the {\OccPT} train set.
Since {\model} outputs part-based features, we made some adjustments to ByteTrack's tracklet management strategy, since it only supports global {\reid} methods outputting a single feature vector per detection.
We therefore compute a tracklet's part-based features as a moving average of the part-based features of the underlying detections.
If some body parts are not visible within a detection that has been matched to a given tracklet, the corresponding part-based features in the tracklet are not updated.
Performance in Table \ref{table:tracking} are reported on the PoseTrack21 \cite{Doering2022} validation set.


\subsection*{Complete Ablations Studies and SOTA Table}
We provide a more comprehensive state-of-the-art comparison and ablation studies in \cref{table:sota_full} and \cref{table:full_ablation_study}.
The complete ablation study includes additional experiments to demonstrate the impact of {\model} components when starting from the SOLIDER \cite{SOLIDER} pre-trained weights.
It also includes an experiment (Exp. 9) to showcase the impact of keypoint prompting on a global {\reid} method, i.e., on a method where the part-based head is removed, and replaced with a simple global average pooling of the encoder output tokens.


\begin{table}[t!]
\centering
\caption{
Comparison of {\model} with SOTA methods.
Results in \textit{Italic} are not provided in the original paper but reproduced by ourselves.
The 1$^{st}$/2$^{nd}$/3$^{rd}$ best scores are indicated with $\textcolor{red}{^{1/2/3}}$.
}
\label{table:sota_full}
\resizebox{0.8\columnwidth}{!}{
\begin{tabular}{|c|cc|cc|cc|cc|cc|}
\hline
\multicolumn{1}{|c|}{\multirow{2}{*}{\makecell[c]{Datasets}}} &
\multicolumn{2}{c|}{\multirow{2}{*}{\makecell[c]{Market-\\1501}}} &
\multicolumn{2}{c|}{\multirow{2}{*}{\makecell[c]{Occluded-\\reID}}} &
\multicolumn{2}{c|}{\multirow{2}{*}{\makecell[c]{Partial-\\reID}}} &
\multicolumn{2}{c|}{\multirow{2}{*}{\makecell[c]{\textbf{Occluded}-\\\textbf{PoseTrack}}}} &
\multicolumn{2}{c|}{\multirow{2}{*}{\makecell[c]{Occluded-\\Duke}}}
\\
 & & & & & & & & & & \\
 \hline
Type &
\multicolumn{2}{c|}{\multirow{1}{*}{\makecell[c]{Holistic}}} &
\multicolumn{8}{c|}{\multirow{1}{*}{\makecell[c]{Occluded}}} \\
\hline
Object Occlusions &
\multicolumn{2}{c|}{} &
\multicolumn{2}{c|}{$\checkmark$} &
\multicolumn{2}{c|}{$\checkmark$} &
\multicolumn{2}{c|}{} &
\multicolumn{2}{c|}{$\checkmark$}
\\
\hline
Person Occlusions &
\multicolumn{2}{c|}{} &
\multicolumn{2}{c|}{} &
\multicolumn{2}{c|}{} &
\multicolumn{2}{c|}{$\checkmark$} &
\multicolumn{2}{c|}{$\checkmark$}
\\
\hline
\hline
Methods & {R-1}&{mAP}&{R-1}&{mAP}&{R-1}&{R-3}&{R-1}&{mAP}&{R-1}&{mAP} \\
\hline
BoT \cite{BoT} &          94.5 & 85.9 & 58.4 & 52.3 & - & - & \textit{78.8} & \textit{69.7} & 51.4 & 44.7 \\
SGAM \cite{SGAM} &       91.4 & 67.3 & - & - & 74.3 & 82.3 & - & - & 55.1 & 35.3 \\
PGFA \cite{PGFA} &       91.2 & 76.8 & - & - & 68.0 & 80.0 & - & - & 51.4 & 37.3 \\
PCB \cite{PCB} &         93.8 & 81.6 & - & - & - & - & \textit{81.7} & \textit{71.2} & 51.2 & 40.8 \\  
MHSA \cite{MHSA-Net} &   94.6 & 84.0 & - & - & 85.7 & 91.3 & - & - & 59.7 & 44.8 \\
VGTri \cite{VGTri}&     -    & -    & 81.0 & 71.0 & 85.7 & 93.7$\textcolor{red}{^{3}}$ & - & - & 62.2 & 46.3 \\
OAMN \cite{OAMN} &        93.2 & 79.8 & - & - & 86.0 & - & - & - & 62.6 & 46.1 \\
HG \cite{HG} &            95.6 & 86.1 & - & - & 74.8 & 87.3 & - & - & 61.4 & 50.5 \\
PVPM \cite{PVPM} &       -    & -    & 66.8 & 59.5 & 78.3 & - & - & - & -    & -    \\ 
HOReID \cite{HOReID} &   94.2 & 84.9 & 80.3 & 70.2 & 85.3 & 91.0 & - & - & 55.1 & 43.8 \\
ISP \cite{ISP} &         95.3 & 88.6 & - & - & - & - & - & - & 62.8 & 52.3 \\
PAT \cite{PAT} &         95.4 & 88.0 & 81.6 & 72.1 & 88.0$\textcolor{red}{^{3}}$ & 92.3 & - & - & 64.5 & 53.6 \\  
PGFL \cite{PGFL-KD} &    95.3 & 87.2 & 80.7 & 70.3 & 85.1 & 90.8 & - & - & 63.0 & 54.1 \\
TRANS \cite{transreid} & 95.2 & 88.9 & - & - & - & - & \textit{83.5} & \textit{73.4} & 66.4 & 59.2 \\  %
SOLIDER \cite{SOLIDER} & \textbf{96.9}$\textcolor{red}{^{1}}$ & \textbf{93.9}$\textcolor{red}{^{1}}$ & - & - & - & - & \textit{84.4} & \textit{76.1}$\textcolor{red}{^{3}}$ & \textit{71.2} & \textit{61.9} \\  
SSGR \cite{OPReID} &      96.1 & 89.3 & 78.5 & 72.9 & - & - & - & - & 69.0 & 57.2 \\
FED \cite{FED} &          95.0 & 86.3 & \textbf{86.3}$\textcolor{red}{^{1}}$ & 79.3$\textcolor{red}{^{3}}$ & 84.6 & - & - & - & 68.1 & 56.4 \\
LDS \cite{LDS} &          95.8 & 90.3 & - & - & - & - & - & - & 64.3 & 55.7 \\
BPBreid \cite{bpbreid} & 95.7 & 89.4 & 82.9 & 75.2 & - & - & \textit{84.9} & \textit{75.5} & 75.1 & 62.5 \\
PFD \cite{PFD} &         95.5 & 89.7 & 83.0 & 81.5$\textcolor{red}{^{2}}$ & - & - & - & - & 69.5 & 61.8 \\
\hline
{\model}\textsubscript{IN} w/o prompt & 95.6 & 88.7 & 83.3 & 78.2 & 81.7 & 86.0 & 85.3 & 75.4 & 75.8 & 63.4 \\
\rowcolor{Pink}
{\model}\textsubscript{IN} & 95.9 & 89.6 & 85.4$\textcolor{red}{^{2 }}$ & 79.1 & 86.0 & 90.0 & \textbf{92.3}$\textcolor{red}{^{1}}$ & \textbf{82.3}$\textcolor{red}{^{1}}$ & 79.8$\textcolor{red}{^{3}}$ & 67.1$\textcolor{red}{^{3}}$ \\
\hline
{\model}\textsubscript{SOL} w/o prompt & 96.6$\textcolor{red}{^{3}}$ & 93.0$\textcolor{red}{^{3}}$ & 80.0 & 78.5 & 90.3$\textcolor{red}{^{2}}$ & 93.7$\textcolor{red}{^{2}}$ & 86.1$\textcolor{red}{^{3}}$ & 75.8 & 82.5$\textcolor{red}{^{2}}$ & 73.3$\textcolor{red}{^{2}}$ \\
\rowcolor{Pink}
{\model}\textsubscript{SOL} & 96.6$\textcolor{red}{^{2}}$ & 93.2$\textcolor{red}{^{2}}$ & 84.8$\textcolor{red}{^{3}}$ & \textbf{82.6}$\textcolor{red}{^{1}}$ & \textbf{90.7}$\textcolor{red}{^{1}}$ & \textbf{94.0}$\textcolor{red}{^{1}}$ & 90.6$\textcolor{red}{^{2}}$ & 81.2$\textcolor{red}{^{2}}$ & \textbf{84.3}$\textcolor{red}{^{1}}$ & \textbf{75.1}$\textcolor{red}{^{1}}$ \\

\hline
\end{tabular}
}
\end{table}

\begin{table}[t!]
\centering
\caption{
Complete ablation study of our proposed KPR architecture. 
PBH stands for ``Part-based Head", meaning we use multiple part-based embeddings with visibility scores instead of a single global embedding (\cref{section:pbh}).
MSF stands for the ``Multi-Scale Features" module added to Swin (\cref{section:encoding}). 
Prompt $\oplus$/$\ominus$ refers to positive/negative prompts (\cref{section:tokenization}).
{\randocc} stands for the ``Batch-wise Inter-Person Occlusion" data augmentation (\cref{section:bipo}). 
SOL stands for the pre-trained weights from SOLIDER \cite{SOLIDER} (\cref{section:implementation}).
}
\label{table:full_ablation_study}
\resizebox{0.85\columnwidth}{!}{
\begin{tabular}{|c|c|c|c|c|c|c|c|c|c|c|}
\hline
\multirow{3}{*}{Id} &\multicolumn{6}{c|}{Main components of {\model}} & \multicolumn{4}{c|}{Dataset} \\ 
\cline{2-11} 
 & \multirow{2}{*}{PBH} & \multirow{2}{*}{MSF} &  \multicolumn{2}{c|}{Prompt} & \multirow{2}{*}{BIPO} & \multirow{2}{*}{SOL} & \multicolumn{2}{c|}{Occluded-PoseTrack} & \multicolumn{2}{c|}{Occluded-Duke} \\
\cline{4-5} \cline{8-11} 
  &  &  & \multicolumn{1}{c|}{$\oplus$} & \multicolumn{1}{c|}{$\ominus$} &  &  & \multicolumn{1}{c|}{R-1} & \multicolumn{1}{c|}{mAP} & \multicolumn{1}{c|}{R-1} & \multicolumn{1}{c|}{mAP} \\ 
\hline
1 &  &  &  &  &  &                                                                  & 84.8                           & 74.7                           & 61.8                       & 52.4 \\  %
2 & \checkmark &  &  &  &  &                                                        & 83.3 (\textcolor{gray}{-1.5})  & 72.2 (\textcolor{gray}{-2.5})  & 67.6 (\textcolor{PositiveGreen}{+5.8}) & 54.2 (\textcolor{PositiveGreen}{+1.8}) \\  %
3 & \checkmark & \checkmark &  &  &  &                                              & 84.8 (\textcolor{PositiveGreen}{+0.0})  & 76.1 (\textcolor{PositiveGreen}{+1.4}) & 68.9 (\textcolor{PositiveGreen}{+7.1}) & 56.8 (\textcolor{PositiveGreen}{+4.4}) \\  %
4 & \checkmark & \checkmark & \checkmark &  &  &                                    & 88.8 (\textcolor{PositiveGreen}{+4.0}) & 80.4 (\textcolor{PositiveGreen}{+5.7}) & 70.8 (\textcolor{PositiveGreen}{+9.0}) & 58.7 (\textcolor{PositiveGreen}{+6.3}) \\  %
5 & \checkmark & \checkmark & \checkmark & \checkmark &  &                          & 90.0 (\textcolor{PositiveGreen}{+5.2}) & 80.7 (\textcolor{PositiveGreen}{+6.0}) & 73.8 (\textcolor{PositiveGreen}{+12.0}) & 62.0 (\textcolor{PositiveGreen}{+9.6}) \\  %
6 &  &  & \checkmark & \checkmark &  &                          & 87.2 (\textcolor{PositiveGreen}{+2.4}) & 78.9 (\textcolor{PositiveGreen}{+4.2}) & 69.0 (\textcolor{PositiveGreen}{+7.2}) & 59.7 (\textcolor{PositiveGreen}{+7.3}) \\  %
7 & \checkmark & \checkmark &  &  & \checkmark &                                    & 86.1 (\textcolor{PositiveGreen}{+1.3}) & 76.7 (\textcolor{PositiveGreen}{+2.0}) & 75.8 (\textcolor{PositiveGreen}{+14.0}) & 63.4 (\textcolor{PositiveGreen}{+11.0}) \\  %
\rowcolor{Pink}
8 & \checkmark & \checkmark & \checkmark & \checkmark & \checkmark &                & \textbf{92.3} (\textcolor{PositiveGreen}{+6.5}) & \textbf{82.3} (\textcolor{PositiveGreen}{+7.6}) & 79.8 (\textcolor{PositiveGreen}{+18.0}) & 67.1 (\textcolor{PositiveGreen}{+14.7}) \\  %
\rowcolor{Pink}
9 & \checkmark & \checkmark & \checkmark & \checkmark & \checkmark & \checkmark     & 90.6 (\textcolor{PositiveGreen}{+5.8}) & 81.2 (\textcolor{PositiveGreen}{+6.5}) & \textbf{84.3} (\textcolor{PositiveGreen}{+22.5}) & \textbf{75.1} (\textcolor{PositiveGreen}{+22.7}) \\  %
\hline
10 &  & \checkmark & \checkmark & \checkmark & \checkmark &                          & 89.9 (\textcolor{PositiveGreen}{+5.1}) & 80.4 (\textcolor{PositiveGreen}{+5.7}) & 72.2 (\textcolor{PositiveGreen}{+10.4}) & 62.0 (\textcolor{PositiveGreen}{+9.6}) \\  %
11 &  &  &  &  &  & \checkmark                                                       & 84.4 (\textcolor{gray}{-0.4}) & 76.1 (\textcolor{PositiveGreen}{+1.4}) & 71.2 (\textcolor{PositiveGreen}{+9.4}) & 61.9 (\textcolor{PositiveGreen}{+9.5}) \\  %
12 & \checkmark & \checkmark &  &  &  & \checkmark                                  & 84.7 (\textcolor{gray}{-0.1}) & 76.6 (\textcolor{PositiveGreen}{+1.9}) & 77.2 (\textcolor{PositiveGreen}{+15.4}) & 68.8 (\textcolor{PositiveGreen}{+16.4}) \\  %
13 & \checkmark & \checkmark & \checkmark & \checkmark &  & \checkmark              & 89.1 (\textcolor{PositiveGreen}{+4.3}) & 79.2 (\textcolor{PositiveGreen}{+4.5}) & 78.1 (\textcolor{PositiveGreen}{+16.3}) & 70.5 (\textcolor{PositiveGreen}{+18.1}) \\  %
\hline
\end{tabular}
}
\end{table}


\subsection*{Qualitative Results}
\begin{figure}[h!]
\centering
\includegraphics[width=0.8\linewidth]{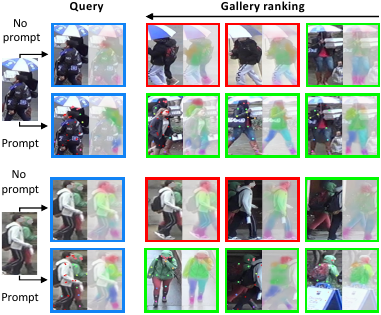}
  \caption{
    Ranked gallery samples for a given query, when the input prompt is disabled/enabled.
    Green/red borders are correct/incorrect matches. 
  }
\label{fig:qualitative_results_supp}
\end{figure}

We provide additional qualitative results in Figure \ref{fig:qualitative_results_supp}.

\subsection*{Additional Implementation Details}
The SOLIDER \cite{SOLIDER} backbone was challenging to fine-tune and we found out 1) the ``\textit{semantic\_weight}" parameter did not have much impact and was set to 0.2 in our experiments, and 2) it was necessary to freeze the keypoint tokenizer for the first 20 epochs. 
The label smoothing \cite{Adaimi2021} regularization rate $\varepsilon$ is set to $0.1$, and the triplet loss margin \cite{triplet} $\alpha$ is set to $0.3$.
The token input and output dimensions $C_i$ and $C_o$ are respectively set to $128$ and $1024$.
All resulting part-based embeddings \{$f_1$, ..., $f_K$\} are further downsampled to an output dimension of $256$ with a simple linear layer followed by a batch normalization and ReLu operation, in order to reduce memory footprint.
Images are first augmented with random cropping and 10 pixels padding, and then with random erasing \cite{random-erasing} at $0.5$ probability, before applying our propose BIPO data augmentation. 
A training batch consists of 64 samples from 16 identities with 4 images each.
The model is trained in an end-to-end fashion for 120 epochs with the SGD optimizer on one NVIDIA RTX8000 GPU.
We employ a cosine annealing learning rate scheduler with 5 epochs warmup.
For {\model}\textsubscript{IN}/{\model}\textsubscript{SOL}, images are resized to a width of $128$ and a height of $256$/$384$, and the learning rate is initialized to $0.008$/$0.0002$.



\end{document}